# Intelligent Railway Foreign Object Detection: A Semi-supervised Convolutional Autoencoder Based Method

Tiange Wang, Zijun Zhang, Fangfang Yang, and Kwok-Leung Tsui

*Abstract*—Automated inspection and detection of foreign objects on railways is important for rail transportation safety as it helps prevent potential accidents and trains derailment. Most existing vision-based approaches focus on the detection of frontal intrusion objects with prior labels, such as categories and locations of the objects. In reality, foreign objects with unknown categories can appear anytime on railway tracks. In this paper, we develop a semi-supervised convolutional autoencoder based framework that only requires railway track images without prior knowledge on the foreign objects in the training process. It consists of three different modules, a bottleneck feature generator as encoder, a photographic image generator as decoder, and a reconstruction discriminator developed via adversarial learning. In the proposed framework, the problem of detecting the presence, location, and shape of foreign objects is addressed by comparing the input and reconstructed images as well as setting thresholds based on reconstruction errors. The proposed method is evaluated through comprehensive studies under different performance criteria. The results show that the proposed method outperforms some well-known benchmarking methods. The proposed framework is useful for data analytics via the train Internet-of-Things (IoT) systems.

## I. Introduction

WITH the expansion of metro systems in intra- and inter-city transportations, an increasing attention has been paid to the reliability and safety of operating railway systems [1]. Railway accidents such as train damages or derails caused by unexpected objects or events are the most important issues concerned by the automatic inspection of railways. Potential foreign objects on or near railway tracks are commonly caused by human beings, such as railway workers and pedestrians, and may threaten the safety of metro transportation [2]. Thus, it is of a great significance to study technologies facilitating the detection of foreign objects.

A large majority of literature relating to foreign objects detection and obstacle detection of railways use infrared and ultrasonic range sensors mounted on the front of trains [3-5]. Although equipping these sensing devices does not require significant changes to the current railway systems, especially to on-ground facilities, their advantages are restricted by distances and weather conditions. Furthermore, special sensors including stereo cameras, radar, and LIDAR are prohibitively expensive and require considerable power in applying them into the railway automatic inspection. The recent advancement of deep learning techniques has driven the development of computer vision based approaches for railway inspections [6]. Most methods presume the scope of types of foreign objects. However, in practice, driving scenes can be complicated in operating railway systems and it is impossible to pre-define a set of objects covering all possibilities. Meanwhile, in captured images, it is quite natural that positive samples containing foreign objects are much less than negative samples which do not contain foreign objects. Therefore, to improve the practical value of foreign object detection via machine learning techniques, a new method well addressing the problem of foreign object detection with insufficient samples of interests needs to be developed.

In this paper, we propose an intelligent railway foreign object detection (RFOD) method based on a convolutional autoencoder (CAE) developed via adversarial training to automate the analytics of railway track images and the detection of various foreign objects. Images considered as inputs of our developed method are captured by two cameras mounted on the train bottom. The RFOD model developed by the proposed method generates a reconstructed image of each raw input via a tailor-made convolutional autoencoder module as well as optimizes the visual quality through a discriminator. To train the proposed model, we design a loss function including a perceptual loss and a pixel loss. The training process of the proposed method is semi-supervised which only takes images without any foreign objects as inputs (normal images) so that the CAE module is trained to generate images describing normal versions of any railway track images. To apply the developed RFOD model, anomaly score measuring the dissimilarity between the input image and its reconstructed image is designed and calculated. Normal images tend to have lower anomaly scores than abnormal images containing at least one foreign object. Additionally, by subtracting the average reconstruction error in training set as a post-processing step, the difference between abnormal images as inputs and their

T. Wang, Z. Zhang, and F. Yang are with the School of Data Science, City University of Hong Kong, Hong Kong. (Corresponding author: Zijun Zhang, zijzhang@cityu.edu.hk).

K.-L Tsui is with Grado Department of Industrial and Systems Engineering, Virginia Polytechnic Institute and State University, Blacksburg, VA, USA.

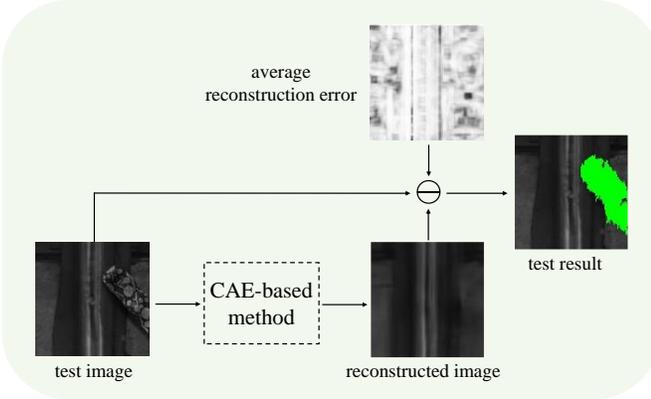

Fig. 1. Test sample illustrating the performance of the proposed RFOD method.

reconstructions can be sharpened to better detect and localize the foreign objects. Test samples for RFOD are displayed in Fig. 1. Since cameras are mounted on the train bottom, foreign objects near the tracks can be well captured. With high authenticity of the reconstructed image, the foreign object is clearly localized and segmented with an instance mask in Figure 1. Based on computational results, the developed method performs robustly in identifying abnormal images and provides accurate segmentation masks for each foreign object. Its advantages against traditional methods are verified via the computational experiments based on the considered image set of railway track.

To summarize, the main contributions of this research are described in the following:
1. We propose a new RFOD method, a semi-supervised CAE-based method, for facilitating the railway inspection automation. The presented method processes railway track images and offers three integrated modules to detect and localize foreign objects. Our proposal is proven to be effective with even limited training data.
2. In addition, we design different constraints on the authenticity of reconstructed images which facilitate the model training. The architecture of the RFOD model is well explored through computational experiments and offered in this paper.

The RFOD model can achieve a high accuracy and perform robustly with regard of various emerging foreign objects, representing its suitability as a practical solution for the RFOD automation.

## II. RELATED WORK

Early studies relating to foreign object detections focused on developing sensors and signal processing technologies. In [7], a multi-sensor system including Lidar and video cameras were implemented to detect frontal obstacles. In [8], a multi-baseline stereo technique was utilized to detect small obstacles at a long range on highways. Similarly, a real-time environment recognition system was developed in [9] to measure the frontal objects within the range of 100m, which was durable to changes of the natural environment. In [10], authors proposed a framework based on infrared sensors as well as GSM and GPS signals to address the obstacle detection and the train tracking. Although sensor-based approaches present potential solutions to foreign object detection tasks, they usually require a relatively clean environment with limited distances and stable weather conditions to compute the disparity between normal and abnormal cases with a high accuracy.

With the advancement of image processing techniques, it has been effective to apply machine visions into numerous real-world applications. The railway monitoring also benefits from the emerging development of visual sensors and machine learning (ML) principles [11, 12]. Compared with sensor-based approaches, ML-based methods process raw images captured by visual sensors and convert the input into rich features by applying advanced ML techniques. Bayesian SegNet was proposed in [13] to estimate label probability and incorporate a threshold to detect unknown objects, which coincide with lower confidence regions in the predicted confidence map. Except for the application of Bayesian networks, there are other methods relying on the approximation of a density function of motion features. For example, optical flow and trajectory analysis were used to exploit non-parametric estimators of unusual event in [14]. Obstacles were detected by applying an image subtraction to corresponding frames in [15, 16] with video data. They aligned two image sequences and provided advantages for detecting unknown objects. These types of work mainly focus on railway scenarios in the video surveillance, which requires a great volume of data leading to more calculations based on temporal and spatial alignments. In practice, these probabilistic model based methods are not easy to train and tend to yield many false positives in irrelevant regions. A simpler way to detect anomalies in images is using distances to classify whether the image is anomalous. One-class classification approaches were trained only on normal data, such as the One Class Support Vector Machines (OC-SVM) [17] and Isolation Forests (IF) [18], which learnt classification boundaries using the normal data, and were widely applied. Nevertheless, these distance-based methods were only studied for the image-level classification. They did not offer the function of detecting anomalous regions on input images.

Deep learning (DL) techniques has progressed tremendously in recent years and their applications on the railway system infrastructure inspection have been recently explored. In [19], an automated railway inspection system was proposed to localize and classify anomalous objects in thermal images captured during night. The framework had three different CNN modules to realize the recognition, localization, and classification sequentially. In [20], a differential feature fusion convolutional neural network (DFF-Net) was introduced to detect traffic obstacles on the railway tracks in the shunting mode. It was composed of two modules, the prior object detection module for generating initial anchor boxes and the differential feature fusion sub-module for enriching the sematic information. These DL-based approaches typically operate under the assumption that all foreign objects encountered in testing have been seen in training, which is obviously unrealistic. More recent vision-based inspection studies focused on reconstructing normal data and subsequently identified anomalies with relatively high reconstruction errors. In [21], an autoencoder (AE) with a semantic segmentation was trained with only images from normal road scenes. It assumed that never-seen-before objects would be decoded poorly in reconstructed image thus an anomaly map was predicted to detect the accurate size and shape of the obstacles. The same principle applies to generative adversarial networks (GAN). In GAN-based approaches [22], given an original distribution $z$ (usually drawn from Gaussian or uniform distribution), the algorithm searched for the latent vector to yield an image that could match the input most closely. In principle, foreign objects from abnormal images can be distinguished using these reconstruction-based approaches. However, they tend to generate the image as a lower-quality version of the input image without explicitly imposing constraints on the authenticity of the reconstructions. Besides, the GAN-based method needs to solve an optimization problem for each example at the inference time to find a latent $z$ such that the difference score between reconstructed image and test image is smallest, which is computationally expensive.

In general, we notice that most of the vision-based anomaly detection in railway studies is trained in a supervised manner, which requires considerably sufficient training samples from both normal and abnormal images. Few unsupervised and semi-supervised methods target on discussing the detection of anomalies in the railway track images. Therefore, we develop the RFOD model, a semi-supervised CAE-based detector, to identify foreign objects near or on tracks by first generating authentic normal versions of abnormal images and next matching them to make a comparison.

## III. METHOD DESCRIPTION

The feasibility of employing generative models to study the anomaly detection has been studied in [23]. In this study, we develop a CAE-based framework composed of three sequential modules, an encoder network $E_\theta$, a decoder network $G_\phi$, and a discriminator model $D_\omega$, to particularly address the automated RFOD based on image data captured by visual sensors. As described in Fig. 2, $E_\theta$ first receives the image $x \in \mathbb{R}^{C \times H \times W}$ as the input (please note that it receives only normal images in training while any images in testing) and projects it into a lower dimensional manifold $z \in \mathbb{R}^d$ with parameters $\theta$. Next, $G_\phi$ produces an reconstruction from $z$ to $\hat{x} \in \mathbb{R}^{C \times H \times W}$, which is a normal version of the input images excluding any foreign objects. Subsequently, $D_\omega$ acts as a binary classifier to determine whether $\hat{x}$ has a high visual quality and can be qualified as an authentic image, just like $x$. Finally, the reconstruction $\hat{x}$ is compared with the input $x$ through the loss function to update model parameters. To apply the developed method to identify and analyze abnormal images, an anomaly score $A(x, \hat{x})$ and a reconstruction error map $C(x, \hat{x})$ are designed to detect the presence of foreign objects. The details of each module, in terms of task, architecture, and training procedure, are described below.

The input of the first module $E_\theta$ is a color image captured by the image acquisition system mounted on the train bottom. Since the raw images collected by the acquisition system are not exactly of the same size, they are resized to a fixed height and width (i.e. $128 \times 128$) in consideration of the model accuracy and computing burden. $E_\theta$ consists of 4 convolutional layers (Conv) downscaling $x$ to smaller feature maps and one fully connected layer (FC) flattening the feature maps to a vector $z \in \mathbb{R}^{512}$, which is also known as the bottleneck features and hypothesized to have a low dimensional representation of $x$. The module of $E_\theta$ is responsible to provide latent features form inputs rather than randomly generated variables, which is more efficient than GAN-based approaches [22]. The architecture of $E_\theta$ is described in **Table 1**. It consists in 4 Conv layers with stride $s = 2$, one FC layer with stride $s = 1$. Most of the layers have kernel size $k = 5$, except for the intermediate Conv_3 layer, which has $k = 4$ but with padding $p = 1$. The size of the feature map is doubled at each layer, starting from 32, arriving to 512 in the bottleneck. For all Conv layers, LeakyReLu with slope $s = 10^{-2}$ is used as the activation function after batch normalization (BN).

With a feature representation $z$, the decoder part $G_\phi$ generates $\hat{x}$ via a symmetrical structure. As described in **Table 1**, $G_\phi$ is composed of 5 transpose convolutional layer (Trans. Conv) to up-sample the vector $z$. The first layer has stride $s =$

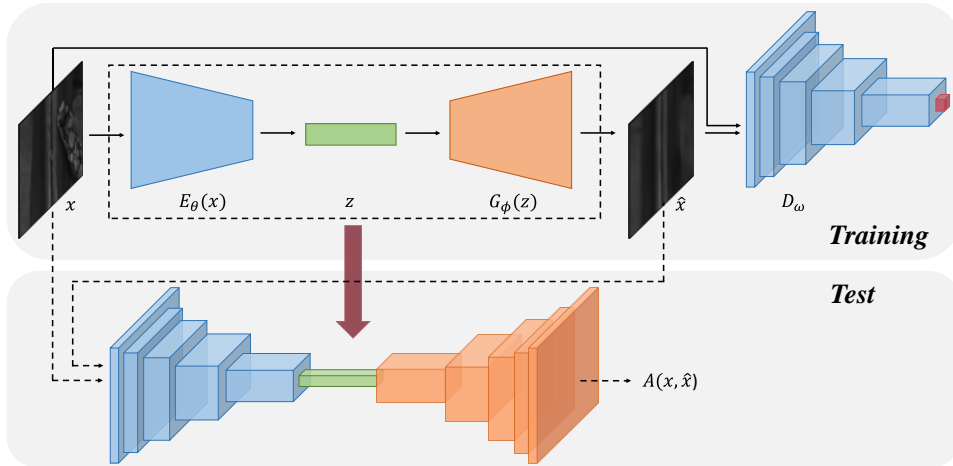

Figure 2. Pipeline of the proposed approach for RFOD on railway tracks.

**Table 1.** Architecture of the proposed framework
(*: padding = 1, †: dropout = 0.3)

| Layer | Kernel | Stride | Filter Units | Output Shape | BN | Activation |
|---|---|---|---|---|---|---|
| $E_\theta(x)$ | | | | | | |
| Input | / | / | / | (128, 128, 3) | / | / |
| Conv_1 | 5×5 | 2 | 32 | (62, 62, 32) | √ | LeakyReLu |
| Conv_2 | 5×5 | 2 | 64 | (29, 29, 64) | √ | LeakyReLu |
| Conv_3* | 4×4 | 2 | 128 | (14, 14, 128) | √ | LeakyReLu |
| Conv_4 | 5×5 | 2 | 256 | (5, 5, 256) | √ | LeakyReLu |
| FC | 5×5 | 1 | 512 | (1, 1, 512) | × | / |
| $G_\phi(z)$ | | | | | | |
| Trans. Conv_1 | 5×5 | 1 | 256 | (5, 5, 256) | √ | ReLu |
| Trans. Conv_2 | 5×5 | 2 | 128 | (14, 14, 128) | √ | ReLu |
| Trans. Conv_3* | 4×4 | 2 | 64 | (29, 29, 64) | √ | ReLu |
| Trans. Conv_4 | 5×5 | 2 | 32 | (62, 62, 32) | √ | ReLu |
| Trans. Conv_5 | 5×5 | 2 | 3 | (128, 128, 3) | √ | Tanh |
| $D_\omega(x, \hat{x})$ | | | | | | |
| Conv_1 | 5×5 | 2 | 32 | (62, 62, 32) | √ | LeakyReLu |
| Conv_2 | 5×5 | 2 | 64 | (29, 29, 64) | √ | LeakyReLu |
| Conv_3*† | 4×4 | 2 | 128 | (14, 14, 128) | × | LeakyReLu |
| Conv_4† | 5×5 | 2 | 256 | (5, 5, 256) | × | LeakyReLu |
| FC | 5×5 | 1 | 1 | (1, 1, 1) | × | Sigmoid |

1 while remaining layers consider $s = 2$. Similar to $E_\theta$, the intermediate Trams. Conv_3 layer has kernel size $k = 4$ and padding equals 1. Starting from 512, the size of the feature map is halved layer-wise and is finally reduced to 32 at the end of $G_\phi$. The final output is still an image with $128 \times 128$ pixels and 3 channels, so that the reconstruction and input lie closely on the manifold. By applying BN into each layer, ReLu activation is used in most of layers. The last layer in $G_\phi$ utilizes Tanh as the activation function to introduce non-linearity in the final output. The combination of $E_\theta$ and $G_\phi$ is referred as an AE architecture, which is trained via a semi-supervised approach. During the training, the $E_\theta$ receives only normal images without foreign objects. However, in testing, $E_\theta$ receives both normal and abnormal images while $G_\phi$ is supposed to output $\hat{x}$ excluding any foreign objects.

The discriminator network $D_\omega$ is a replica of $E_\theta$ but with two differences. One is removing BN in the last two Conv layers and applying the dropout regularization with a drop probability $p = 0.2$ (as displayed in Table 1). Another is that the sigmoid function is applied at the end of $D_\omega$ to discriminate $x$ and $\hat{x}$, which has been discussed in [24]. The $D_\omega$ is designed to drive the AE structure to stably produce images with a high authenticity. The two parts repeatedly compete against each other through adversarial training. The result of such competition is that $\hat{x}$ is as similar as possible to $x$ so that $D_\omega$ can no longer distinguish them.

The training loss in our proposal has two parts, the generator loss $\mathcal{L}_{E,G}$ and the discriminator loss $\mathcal{L}_D$. Since the architecture of $E_\theta$ and $G_\phi$ aims to reconstruct the input image, we adopt $\mathcal{L}_{pixel}$, the commonly used mean absolute error, as one of the loss terms in $\mathcal{L}_{E,G}$. We use $L_1$ distance in $\mathcal{L}_{pixel}$ rather than $L_2$ as the former one encourages less blurring. In our method, $G_\phi$ is tasked not only to fool $D_\omega$ but also to be near the ground truth output in an $L_1$ sense. In addition, perceptual loss $\mathcal{L}_{perceptual}$ [25], which computes the $L_2$ distance between the encoded feature representation of $x$ and $\hat{x}$, is also considered to improve the authenticity of $\hat{x}$. Overall, the loss function of generator $\mathcal{L}_{E,G}$ is defined as follows:

$$\mathcal{L}_{E,G} = \mathcal{L}_{perceptual} + \mathcal{L}_{pixel} =$$
$$\sum \mathbb{E}_{x,\hat{x}} \|f_i(x) - f_i(\hat{x})\|_2 + \mathbb{E}_{x,\hat{x}} \|x - \hat{x}\|_1, \quad (1)$$

where $f_i(\cdot)$ denotes the features from $i^{th}$ LeakyReLU layer of $E_\theta$ and. $\mathcal{L}_{perceptual}$ encourages the output image $\hat{x}$ to be perceptually similar to the actual image but does not force them to match exactly. Usually, $\mathcal{L}_{pixel}$ itself does not capture perceptual differences between generated and real images because the image might look similar in perspective but with different per-pixel values. Previous works, such as [22] and [23], apply the difference between bottleneck features where $\mathcal{L}_z = \mathbb{E}_{x,\hat{x}} \|z - \hat{z}\|_2$ as one of loss terms while the remaining intermediate feature representations extracted from $E_\theta$ are ignored. We utilize richer features instead of the single bottleneck feature to provide stricter constraints in the training procedure. As for the discriminator loss $\mathcal{L}_D$, it follows the classical discriminator GAN loss, which derives from the cross-entropy between the real and reconstructed distributions:

$$\mathcal{L}_D = \mathbb{E}_x[\log(D_\omega(x))] + \mathbb{E}_z[\log(1 - D_\omega(G_\phi(z)))], \quad (2)$$

where $\mathbb{E}_x$ is the expected discriminant value over real inputs, and $\mathbb{E}_z$ is the expected discriminant value over their reconstructions. In the standard adversarial training, the generator is continuously updated to fool the discriminator while the primary goal of the discriminator is to learn the way of identifying those reconstructions progressively. Thus, when the input of $D_\omega$ is $\hat{x}$, $D_\omega(G_\phi(z))$ should be estimated as small as possible. In other words, $\mathbb{E}_z$ provides the ability to visually assess the quality of $\hat{x}$. The overall objective function consisting of the generator loss $\mathcal{L}_{E,G}$ and the discriminator loss $\mathcal{L}_D$ balance the training on $D_\omega$ while also improve the training on $E_\theta$ and $G_\phi$. The pseudo code of training the RFOD model is presented in **Algorithm 1**. Model parameters $\theta, \phi, \omega$ are updated using AdamW optimizer via the adversarial training of $E_\theta$, $G_\phi$ and $D_\omega$.

In Fig. 2, an anomaly score $A(x, \hat{x})$ is defined to express the fit of $\hat{x}$ to $x$ in the testing. Its computation directly depends on the loss function:

$$A(x, \hat{x}) = \mathbb{E}_{x,\hat{x}} \|f(x) - f(\hat{x})\|_1, \quad (3)$$

where $f(\cdot)$ denotes the last activation layer of $E_\theta$. Instead of using $L_2$ distance, the $L_1$ reconstruction error is applied at the last updating iteration in training to compute the anomaly score. The model is supposed to yield a large $A(x, \hat{x})$ for abnormal

**Algorithm 1** Adversarial training of the proposed RFOD method

**Input:** Sets of real input images $x_i \in X_{train}$, max number of iterations $T$, number of $D_\omega$ updates per epoch ($K_D$), number of $E_\theta$ and $G_\phi$ update per epoch ($K_{AE}$).
**Output:** $E_\theta$, $G_\phi$, and $D_\omega$.

1. Initialize $E_\theta$, $G_\phi$, and $D_\omega$.
2. **for** epoch $i = 1, \dots, T$ **do**
3.    **for** $k = 1, \dots, K_D$ **do**
4.       Sample a mini-batch of training images $x_i$.
5.       $z_i = E_\theta(x_i)$ training images $x_i$.
6.       Update $\omega$ by taking AdamW optimizer on the mini-batch loss $\mathcal{L}_D$ in (2)
7.    **end**
8.    **for** $k = 1, \dots, K_{AE}$ **do**
9.       $\hat{x}_i = G_\phi(E_\theta(x_i))$.
10.      Update $\theta$ and $\phi$ by taking AdamW optimizer on the mini-batch loss $\mathcal{L}_{E,G}$ in (1).
11.    **end**
12. **end**

**Algorithm 2** RFOD in testing.

**Input**: Real input image $x_i \in X_{test}$, $E_\theta$, $G_\phi$, average reconstruction error map $C$ of training set.
**Output**: anomaly score $A(x_i, \hat{x}_i)$.

1. $z_i = E_\theta(x_i)$
2. $\hat{x}_i = G_\phi(z_i)$
3. $A(x_i, \hat{x}_i) = \|f(x_i) - f(\hat{x}_i)\|_1$
4. $M = x_i - \hat{x}_i - C$
5. return $A(x_i, \hat{x}_i)$, $M$

image because it only learns the manifold from normal samples, whereas a small $A(x, \hat{x})$ means that a similar image was already seen during training. Besides, a reconstruction error map $C \in \mathbb{R}^{3 \times 128 \times 128}$ is also calculated as an average $L_1$ loss based on the training set:

$$C(x, \hat{x}) = \frac{1}{N} \sum \|x_i - \hat{x}_i\|_1, \quad (4)$$

where $N$ denotes the number of samples in the training set. This post-processing is applied to filter out small residuals in testing images. Subtracting the $C(x, \hat{x})$, which represents the visual saliency from training samples, helps to better localize the foreign objects. The testing procedure for the developed RFOD model in is described in **Algorithm 2**. The foreign object detection on the image-level only relies on $A(x, \hat{x})$ while the detection on pixel-level depends on the difference map $M$, which is the result of post-processing.

## IV. Computational Experiments

In this section, the dataset used in computational experiments is first introduced. Next, we report the training setup of the whole pipeline and the metrics exploited to assess the detection capability of our method. Finally, results are compared with a set of benchmarking methods, such as two distance-based methods OC-SVM [17] and IF [18] as well as two reconstruction-based methods AE [21] and GAN [22].

### A. Dataset

In our experiment, the image dataset is collected by the Hong Kong Metro Corporation (MTR) through a vision system including two cameras and 6 LED lights as shown in Fig. 3. RGB images are taken vertically towards the track rather than from the bottom of the train, which describes that our targeted RFOD problem is different from previous ones [15, 16, 20, 21]. As displayed in Fig. 1, the foreign object is near the railway track and might not be fully captured. In this image dataset, total 6656 normal images are utilized in training. In testing, we consider additional 579 images, in which 316 of them are of abnormal class containing various foreign objects and 263 of them are normal. Different categories of foreign objects are considered, such as phones, bags, newspapers, etc. The shorter side of raw input images have at least 200 and at most 321 pixels while their longer side has at most 750 pixels. Each of the test images contains at most one foreign object with its image-wise annotations. These annotations are only used for statistical evaluations and are never fed into the network, neither during training nor in the evaluation phase. For the evaluation of the detection performance, a positive label will be assigned to an abnormal image.

The proposed RFOD model is trained via a semi-supervised manner based on normal images only. Considering the scarcity of anomalies, the training volume is set to be much larger than the test volume. Additionally, to better demonstrate the capability of the developed model on identifying the abnormal examples, the number of abnormal images is slightly larger than normal images in testing. To the best of our knowledge, this research is a pioneering work on detection tasks in a railway scenario using images collected by visual sensors mounted on trains.

### B. Training Setup

The proposed networks are optimized based on ($\mathcal{L}_D, \mathcal{L}_{E,G}$), using AdamW with momentums $\beta 1 = 0.5, \beta 2 = 0.9$, and an initial learning rate $lr = 2e - 4$. Experiments are conducted with using three NVIDIA GeForce RTX 2080 GPUs. Each GPU is assigned to process 8 images and the batch size is thus 24. The whole training schedule is empirically set to 100 epochs to yield optimal results. In the proposed method, $G_\phi$ and the AE part are locked into a fierce competition to improve themselves, and thus both are trained in alternating steps. All weights adopt the Xavier initializer for the initialization. The gradient descent starts from $D_\omega$. To avoid the situation where $D_\omega$ wins early in the competition with $G_\phi$ and the gradient of $G_\phi$ vanishes, the iteration of $D_\omega$ is set to one for each iteration of $G_\phi$ and $E_\theta$. In testing, we run the whole pipeline in exactly the same manner as during the training phase. When the batch size is set to 1, the instance normalization is applied instead of the BN.

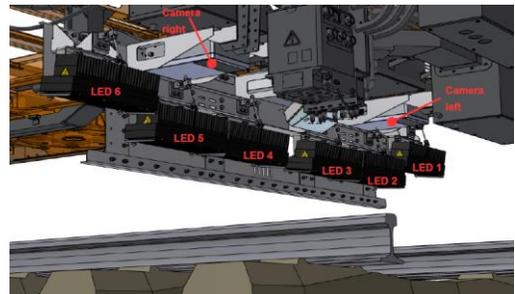

Fig. 3. Location of lights and cameras in the image acquisition system.

The adversarial training is based on the approach described in Section 3 for a consistent comparison. As such, we aim to show the superiority of the proposed architecture regardless of using any tricks to improve the training of $D_\omega$. In addition, the comparison against traditional distance- and reconstruction-based methods is provided to show the advantage of the proposed RFOD framework.

*C. Evaluation Metrics*

The accuracy of our approach is quantitatively evaluated at a level of images. A set of common metrics based on anomaly score $A(x, \hat{x})$ are exploited, such as the area under receiver operating characteristic curve (AUROC) and area under precision-recall curve (AUPRC) as well as the equal error rate (EER). AUROC and AUPRC are defined as follows:

$$AUROC = \sum (f_{n+1} - f_n) t_{interp}(f_{n+1}), \quad (5)$$

$$AUPRC = \sum (r_{n+1} - r_n) p_{interp}(r_{n+1}), \quad (6)$$

where $t(f)$ denotes the relation between the true positive rate (TPR) and false positive rate (FPR) while $p(r)$ denotes the relation between the precision and recall. AUROC measures the ability to discriminate between positive samples (abnormal cases in this paper) and negative samples (normal cases). The ROC curve shows the trade-off between TPR and FPR across different decision thresholds. By using ground-truth labels and $A(x, \hat{x})$ of the test set, TPR and FPR are computed as follows:

$$TPR = \frac{TP}{TP + FN}, \quad (7)$$

$$FPR = \frac{FP}{FP + TN}, \quad (8)$$

where $TP$ is the number of true positives, $FP$ is the number of false positives, and $TN$ is the number of true negatives. Similarly, the operating point (recall, precision) is defined as follows to visualize the PR curve and further compute AUPRC, which is also denoted as the average precision (AP). The recall is intuitively the ability of the classifier to find all positive samples, and the precision is the ability of the classifier that do not label a negative sample as positive. Another metric of F1 score is defined as the harmonic mean of precision and recall:

$$Recall = \frac{TP}{TP + FN} = TPR, \quad (9)$$

$$Precision = \frac{TP}{TP + FP}, \quad (10)$$

$$F1\ score = \frac{2 \times Precision \times Recall}{Precision + Recall}, \quad (11)$$

AUPRC is another useful performance metric for imbalanced data. In our detection task, we pay more attention to properly classifying railway images containing foreign objects. Thus, a higher AUPRC indicates that the model can recognize most of positive examples without mistakenly recognizing negative examples as positive. The baseline for AUROC is always going to be 0.5. Yet, for AUPRC, the baseline is equal to the fraction of positives, which is 0.546 in the test set. To get smother ROC and PR curves, thresholds are obtained after a de-duplication and descending process of A(x, x̂). This allows the ROC curve to start from the bottom left and end at the upper right, whereas opposed to the PR curve. EER is the intersection point between a straight line joining (1, 0) and (0, 1) and the ROC curve. The value of EER indicates the optimization of the trade-off between false positives and false negatives. It is a mathematical way of scoring out errors and error margins in terms of false data, which should be more reasonable than the accuracy metric. Furthermore, the precision and recall at the optimal EER point on the ROC curves are also reported in our experiments. If not specified, all metrics, except for EER, are reported as percentages to provide a fair comparison.

*D. Results and Analysis*

In this section, we conduct the experimental evaluation of the RFOD model. The unified architecture depicted in Fig. 2 is empirically determined in a manual iterative architecture search. We report the results of the whole pipeline in terms of classification accuracy and localization at inference time. The unified architecture of $E_\theta$ listed in Table 1 is then used to train reconstruction-based models coming from [21] and [22] for fair comparison.

Fig. 4 displays the visualization of $\hat{x}$ from the last training iteration. It compares the reconstruction effect on normal training set, normal test set, and abnormal test set, respectively. Based on samples from training set, the output of $G_\phi$ look the same as the real inputs. Even the small components, such as the bolt and clip, are well reflected in the reconstructions. For the normal test set, some of details in the image may not be completely simulated. Yet, the rails are clearly distinguishable. As for the anomalous images in test set, the abnormal area where the foreign object lies on the track becomes vague in $\hat{x}$. Since the detector only learns the manifold from normal samples, all the foreign objects cannot be well simulated through $G_\phi$, which returns to the vague areas and helps to identify the anomalies in testing. The last row of (**c**) shows testing results containing the pixel-level masks and bounding boxes of foreign objects by comparing images from the first row and the second row. Basically, all the foreign objects from the test samples are successfully detected. By applying an error map from (5), most of the normal area, such as rail and bolt, is excluded in the mask and bounding box.

Fig. 5 shows the kernel density estimate of the three sets. In (**a**), $D(G(z))$ is the discriminant value on the reconstructed samples. The mean value of the estimated density on three sets are all close to 0.5. Compared with the test set, the density of training set is more concentrated on mean value. Since the discriminator $D_\omega$ acts as a binary classifier and has sigmoid function at the end of it, the mean value of the estimated density indicates a high probability that $D_\omega$ cannot recognize $\hat{x}$ from $x$. The $\hat{x}$ owns a high authenticity to fool the discriminator, which demonstrates that the generator $G_\phi$ is able to produce realistic images robustly with bottleneck features output from the encoder $E_\theta$. $A(x, G(z))$ is the anomaly score on encoded features from input $x$ and generated $\hat{x}$ as defined in (3). Before the estimation, the score is scaled between [0, 1]. The normal training set and normal test set have the significantly smaller mean value as well as smaller variation than the abnormal test set in terms of the density estimate. Additionally, the wider range of the anomaly score on anomalous set suggests that the proposed model can effectively figure out abnormal images with significantly larger anomaly score.

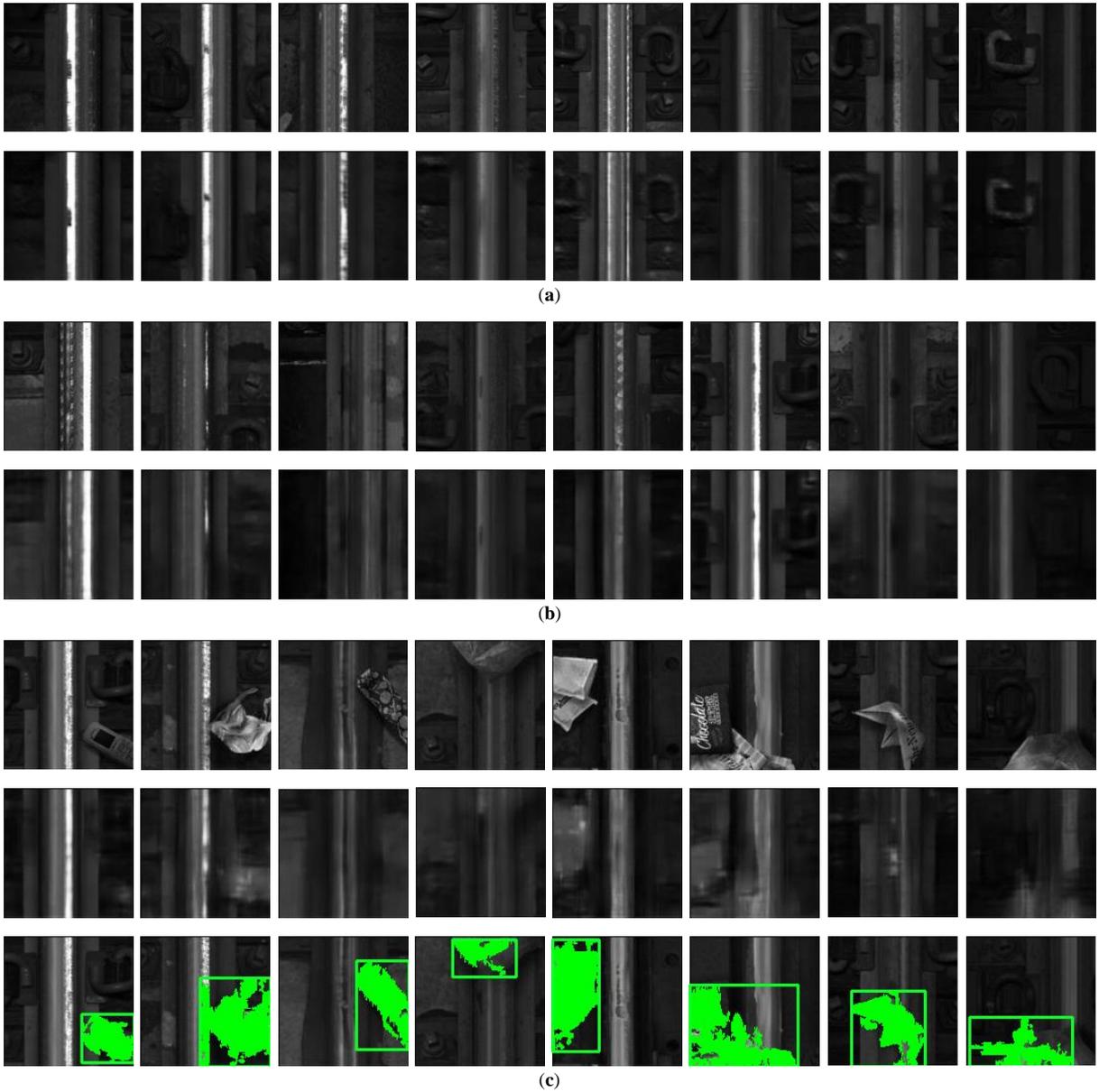

Figure 4. Samples of $x$ and $\hat{x}$. (**a**) All samples are from training set. The first row is real training image $x$, and the second row represents the reconstructed image $\hat{x}$. They look quite similar after the whole training schedule. (**b**) Images from normal test set. The first row is real training images, and the second row represents their reconstructions. The authenticity of reconstructions is not as good as in the training set. (**c**) Images from abnormal test set, which contain different foreign objects. In $\hat{x}$, the abnormal area is quite vague especially when the object lies on the track. The last row displays the inference result of foreign objects including pixel-level masks and bunding boxes.

Besides the visualization of discriminant scores and anomaly scores, we repeated our experiments with and without $\mathcal{L}_{pixel}$. The effect of different loss terms mentioned in Section 3 including $\mathcal{L}_{perceptual}$ and $\mathcal{L}_z$ is also explored. As shown in Fig. 6, the ROC curve and PR curve are plotted using the anomaly scores under different losses. In Fig. 6 (**a**), $\mathcal{L}_{perceptual}$ leads to a higher AUROC than $\mathcal{L}_z$, which indicates that the intermediate feature representations in $E_\theta$ and $D_\omega$ can better depict the similarities between images than the bottleneck features. The points on the two curves are optimal EER points between false positives and false negatives. $\mathcal{L}_{perceptual}$ also has a smaller EER and FPR than $\mathcal{L}_z$. The numbers in the brackets represent AUROC and EER, respectively. With a pixel loss between $x$ and $\hat{x}$ added, both of their accuracies are improved as shown in Fig. 6 (**b**). Similarly, according to PR curves in Fig. 6 (**c-d**), the AUPRC values are generally higher than AUROC because there are more positive samples than negative samples in our test set. Since the primary objective of $E_\theta$ and $D_\omega$ is to generate images as similar as possible to the input images, the $\mathcal{L}_{pixel}$ is the most direct way to reflect the similarity of images. The combination of $\mathcal{L}_{pixel}$ and $\mathcal{L}_{perceptual}$ is used as the generator loss in remaining experiments.

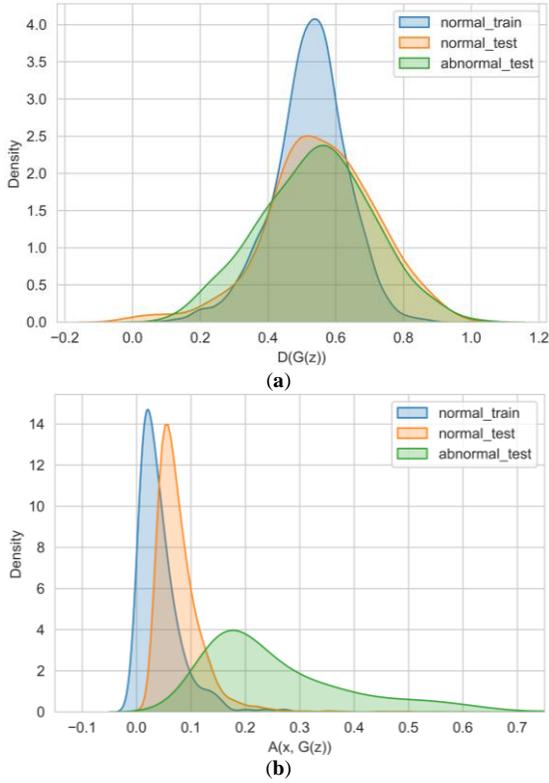

Fig. 5. (**a**) Kernel density estimate of discriminant value on $\hat{x}$. (**b**) Anomaly score between input $x$ and $\hat{x}$.

Then we evaluate our proposed anomaly score and compare it to other reconstruction-based criteria including L1 and L2 reconstruction errors, the bottleneck feature score, as well as the encoded feature score. They are defined as follows. The $x$ is the input sample and $\hat{x} = G_\phi(E_\theta(x))$ is the reconstruction through our proposed method. The $z$ and $\hat{z}$ are bottleneck features extracted through $E_\theta$ from $x$ and $\hat{x}$, respectively. The $f(\cdot)$ denotes the last activations layer from $E_\theta$:

$$L_1(x, \hat{x}) = \|x - \hat{x}\|_1, \qquad (12)$$

$$L_2(x, \hat{x}) = \|x - \hat{x}\|_2, \qquad (13)$$

$$A_{bottle}(x, \hat{x}) = \|E_\theta(x) - E_\theta(\hat{x})\|_1, \qquad (14)$$

$$A_{enc}(x, \hat{x}) = \|f(x) - f(\hat{x})\|_1, \qquad (15)$$

As observed in Table 2, the encoded anomaly score $A_{enc}(x, \hat{x})$ is more suited for the RFOD method as it owns the highest AUROC of 85.66% and AUPRC of 87.72%. Therefore, $A_{enc}(x, \hat{x})$ is applied in our proposal and other computational experiments. However, for the optimized error rate, the bottleneck score $A_{bottle}(x, \hat{x})$ has lower error margins in terms of false classifications, which indicates that $A_{bottle}(x, \hat{x})$ has a better ability to balance the recognition of false positives and false negatives.

Finally, the proposal is compared against several common-used methods for anomaly detections including OC-SVM, IF, AE and traditional GAN. OC-SVM is a classic kernel method learning a decision boundary around normal examples. The negative examples are absent during its training, which is considered as a semi-supervised learning approach as well. IF is also a classic machine learning technique but isolating anomalies instead of modeling the distribution of normal data. It proceeds by first building trees using randomly selected split values across randomly chosen features. Then, the anomaly score is defined to be the average path length from a particular sample to the root. They are both distance-based methods. We use default parameters provided by the scikit-learn package for OC-SVM and IF approaches in the computational experiments. The AE and traditional GAN are reconstruction-based classifiers adopting the same sub-network architecture as the proposed CAE architecture, i.e., the AE applies the same $E_\theta$ and $G_\phi$ to reconstruct images and the GAN utilizes the architecture of $G_\phi$ and $D_\omega$ to measure the quality of the data reconstruction. The comparison between AE, GAN, and RFOD performs an ablation study where different parts of the proposal is removed. Table 3 summarizes the performance of these methods. The proposed RFOD method outperforms the state-of-the-art anomaly detection methods on the railway image dataset in terms of 8.13% higher AUROC and 6.30% higher AUPRC. Besides, the comparison with GAN in Table 3 demonstrates that the latent features of $z$ in $E_\theta$ helps to reconstruct images compared to the random vector without $E_\theta$. Similarly, the discriminator component $D_\omega$ improves the distinguish between $x$ and $\hat{x}$ in our proposal when compared to AE. Furthermore, if we compare the AE and GAN models, the former owns higher AUROC and AUPRC as well as a smaller EER, which indicates that the authenticity of the $\hat{x}$ can be improved by providing a more robust $z$ and the role of discriminator can be replaced with the constraints between $x$ and generated $\hat{x}$.

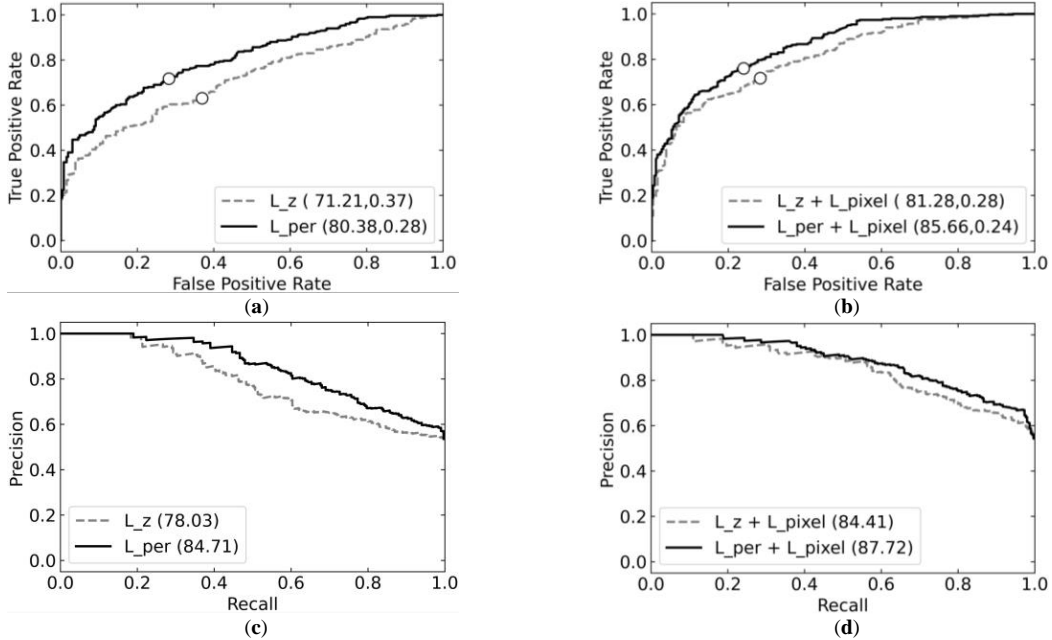

Fig. 6. (**a**) ROC curves of $\mathcal{L}_z$ and $\mathcal{L}_{perceptual}$. The points on the curves are EER. $\mathcal{L}_{perceptual}$ has a higher AUROC as well as a smaller EER (as shown in the brackets). (**b**) $\mathcal{L}_{pixel}$ is added to them. It improves the AUROC of both loss terms. (**c-d**) PR curves of $\mathcal{L}_z$ and $\mathcal{L}_{perceptual}$. The combination of $\mathcal{L}_{perceptual}$ and $\mathcal{L}_{pixel}$ still owns a better performance on AUPRC. Both are expressed as percentages.

**Table 2.** Performance on different score methods.

| $A(x,\hat{x})$ | AUROC | AUPRC | EER | Precision | Recall | F1 score |
|---|---|---|---|---|---|---|
| $L_1$ | 83.07 | 84.83 | 0.27 | 75.84 | 72.70 | 74.24 |
| $L_2$ | 82.61 | 85.14 | 0.26 | 76.90 | 73.82 | 75.33 |
| $A_{bottle}$ | 84.92 | 87.36 | **0.23** | **79.62** | **76.44** | **78.00** |
| $A_{enc}$ | **85.66** | **87.72** | 0.24 | 79.23 | 76.04 | 77.60 |

**Table 3.** Performance on different approaches. We use default parameters provided by the scikit-learn package for OC-SVM and IF. AE and GAN adopt the same sub-network architecture as our proposal. All of them are trained for 100 epochs.

| Model | AUROC | AUPRC | EER | Precision | Recall | F1 score |
|---|---|---|---|---|---|---|
| OC-SVM | 67.49 | 73.03 | 0.38 | 72.98 | 62.06 | 67.08 |
| IF | 65.28 | 68.74 | 0.41 | 60.25 | 59.13 | 59.68 |
| AE | 77.53 | 81.42 | 0.29 | 73.85 | 71.33 | 72.57 |
| GAN | 62.73 | 63.00 | 0.39 | 65.21 | 60.58 | 62.81 |
| **Our proposal** | **85.66** | **87.72** | **0.24** | **79.26** | **76.04** | **77.62** |

## V. Conclusion

We presented a semi-supervised CAE-based method for automating the RFOD, which was composed of an encoder module, a generator, and a discriminator. In the presented method, images were reconstructed by mapping the extracted bottleneck latent features reversely via a symmetric structure. The reconstructed image and the original image were then fed into the discriminator module to assess its authenticity. Via adversarial training between modules, the authenticity of reconstructions was improved to help with the recognition of abnormal images. In the test, the discrepancy between input image and reconstructed image was evaluated through anomaly scores. With the well-trained encoder module, abnormal images containing foreign objects tend to have higher anomaly scores than normal images. Finally, the localization and shape of foreign objects were provided through the dissimilarity between inputs and their reconstructions as well as subtracting the average reconstruction error obtained based on the training set.

In computational experiments, the detection performance of the CAE-based method in terms of different authenticity constraints as well as different test criteria were demonstrated. Moreover, we compared the developed RFOD model with distance- and reconstruction-based benchmarks. It outperformed the benchmarking methods with at least 8.13% higher AUROC and 6.30% higher AUPRC on the railway track dataset.

Our future work on railway inspections will take advantages of the deep learning based detection method from two aspects. First, the potential of the developed CAE-based model on finer inspection tasks, such as surface defect detection, needs to be further explored. Secondly, exploiting the use of temporal image data as well as the segmentation label will be considered to improve the accuracy and the reliability of automated railway inspections.

## References

[1] O. Jo, Y.-K. Kim, J. Kim, Internet of things for smart railway: feasibility and applications, IEEE Internet of Things Journal 5(2) (2017) 482-490.

[2] C. Tastimur, M. Karakose, E. Akin, Image processing based level crossing detection and foreign objects recognition approach in railways, International Journal of Applied Mathematics Electronics and Computers (Special Issue-1) (2017) 19-23.

[3] J.J. García, J. Ureña, A. Hernandez, M. Mazo, J.A. Jiménez, F.J. Álvarez, C. De Marziani, A. Jiménez, M.J. Díaz, C. Losada, Efficient multisensory barrier for obstacle detection on railways, IEEE Transactions on intelligent transportation systems 11(3) (2010) 702-713.

[4] V. Agarwal, N.V. Murali, C. Chandramouli, A cost-effective ultrasonic sensor-based driver-assistance system for congested traffic conditions, IEEE transactions on intelligent transportation systems 10(3) (2009) 486-498.

[5] Y.B. Jinila, A novel approach on obstacle detection and automatic braking system in railways, International Journal of Sciences & Applied Research (IJSAR) (2018).
[6] S. Liu, Q. Wang, Y. Luo, A review of applications of visual inspection technology based on image processing in the railway industry, Transportation Safety and Environment 1(3) (2019) 185-204.
[7] S. Mockel, F. Scherer, P.F. Schuster, Multi-sensor obstacle detection on railway tracks, IEEE IV2003 Intelligent Vehicles Symposium. Proceedings (Cat. No. 03TH8683), IEEE, 2003, pp. 42-46.
[8] T. Williamson, C. Thorpe, Detection of small obstacles at long range using multibaseline stereo, Proceedings of the 1998 IEEE International Conference on Intelligent Vehicles, Citeseer, 1998.
[9] N. TOKUDOME, S. AYUKAWA, S. Ninomiya, S. ENOKIDA, T. NISHIDA, Development of Real-time Environment Recognition System using LiDAR for Autonomous Driving1, International Conference on ICT Robotics, 2017, pp. 25-26.
[10] N.S. Punekar, A.A. Raut, Improving railway safety with obstacle detection and tracking system using GPS-GSM model, International Journal of Scientific & Engineering Research 4(8) (2013) 282-288.
[11] H. Salmane, L. Khoudour, Y. Ruichek, A video-analysis-based railway–road safety system for detecting hazard situations at level crossings, IEEE transactions on intelligent transportation systems 16(2) (2015) 596-609.
[12] A.K. Singh, A. Swarup, A. Agarwal, D. Singh, Vision based rail track extraction and monitoring through drone imagery, ICT Express 5(4) (2019) 250-255.
[13] A. Kendall, V. Badrinarayanan, R. Cipolla, Bayesian segnet: Model uncertainty in deep convolutional encoder-decoder architectures for scene understanding, arXiv preprint arXiv:1511.02680 (2015).
[14] A. Adam, E. Rivlin, I. Shimshoni, D. Reinitz, Robust real-time unusual event detection using multiple fixed-location monitors, IEEE transactions on pattern analysis and machine intelligence 30(3) (2008) 555-560.
[15] H. Mukojima, D. Deguchi, Y. Kawanishi, I. Ide, H. Murase, M. Ukai, N. Nagamine, R. Nakasone, Moving camera background-subtraction for obstacle detection on railway tracks, 2016 IEEE international conference on image processing (ICIP), IEEE, 2016, pp. 3967-3971.
[16] R. NAKASONE, N. NAGAMINE, M. UKAI, H. MUKOJIMA, D. DEGUCHI, H. MURASE, Frontal obstacle detection using background subtraction and frame registration, Quarterly Report of Rtri 58(4) (2017) 298-302.
[17] B. Schölkopf, R.C. Williamson, A.J. Smola, J. Shawe-Taylor, J.C. Platt, Support vector method for novelty detection, NIPS, Citeseer, 1999, pp. 582-588.
[18] F.T. Liu, K.M. Ting, Z.-H. Zhou, Isolation forest, 2008 eighth ieee international conference on data mining, IEEE, 2008, pp. 413-422.
[19] R. Gasparini, A. D'Eusanio, G. Borghi, S. Pini, S. Giuseppe, S. Calderara, F. Eugenio, R. Cucchiara, Anomaly Detection, Localization and Classification for Railway Inspection, 25th International Conference of Pattern Recognition, 2020.
[20] T. Ye, X. Zhang, Y. Zhang, J. Liu, Railway Traffic Object Detection Using Differential Feature Fusion Convolution Neural Network, IEEE Transactions on Intelligent Transportation Systems (2020).
[21] T. Ohgushi, K. Horiguchi, M. Yamanaka, Road Obstacle Detection Method Based on an Autoencoder with Semantic Segmentation, Proceedings of the Asian Conference on Computer Vision, 2020.
[22] T. Schlegl, P. Seeböck, S.M. Waldstein, U. Schmidt-Erfurth, G. Langs, Unsupervised anomaly detection with generative adversarial networks to guide marker discovery, International conference on information processing in medical imaging, Springer, 2017, pp. 146-157.
[23] S. Akcay, A. Atapour-Abarghouei, T.P. Breckon, Ganomaly: Semi-supervised anomaly detection via adversarial training, Asian conference on computer vision, Springer, 2018, pp. 622-637.
[24] I.J. Goodfellow, J. Pouget-Abadie, M. Mirza, B. Xu, D. Warde-Farley, S. Ozair, A. Courville, Y. Bengio, Generative adversarial networks, arXiv preprint arXiv:1406.2661 (2014).
[25] Q. Chen, V. Koltun, Photographic image synthesis with cascaded refinement networks, Proceedings of the IEEE international conference on computer vision, 2017, pp. 1511-1520.